\documentclass[conference]{IEEEtranSPMB}

\IEEEoverridecommandlockouts
\ifCLASSINFOpdf
\else
\fi
\usepackage{amsmath,graphicx,soul,subcaption,balance}
\usepackage{balance}
\usepackage[numbers,sort&compress]{natbib}

\usepackage{parskip}
\usepackage{float} 

\usepackage{mathptmx}

\usepackage{multirow} 

\usepackage{color}
\usepackage{soul,mathrsfs}
\usepackage[utf8]{inputenc}
\usepackage[T1]{fontenc}
\usepackage{array}
\usepackage[hyphens, spaces, obeyspaces]{url}

\usepackage[papersize={8.5in,11in}, left=1in, right=1in, top=1in, bottom=1in]{geometry}
\newcolumntype{L}[1]{>{\raggedright\let\newline\\\arraybackslash\hspace{0pt}}m{#1}}
\newcolumntype{C}[1]{>{\centering\let\newline\\\arraybackslash\hspace{0pt}}m{#1}}
\newcolumntype{R}[1]{>{\raggedleft\let\newline\\\arraybackslash\hspace{0pt}}m{#1}}

\usepackage[singlelinecheck=false,justification=centering]{caption}
\DeclareCaptionLabelFormat{mylabel}{#1 #2.\hspace{0.7ex}}
\renewcommand{\thetable}{\arabic{table}}

\usepackage{xcolor,colortbl}
\definecolor{light-gray}{gray}{0.83}
\newcommand{\GREY}{\cellcolor{light-gray}\bf} 

\newcommand{\spmbtitlefont}{\fontsize{11.0pt}{11.00pt}\selectfont\bf\vspace{0.7em}}
\newcommand{\spmbauthorfont}{\fontsize{11.0pt}{11.0pt}\selectfont\vspace{0em}}
\newcommand{\subparagraph}{}
\usepackage[compact]{titlesec}
\titleformat{\section}
   {\center\normalfont\sc}{\thesection.}{0.7ex}{}
\titlespacing{\section}{0pt}{2ex}{1.5ex}
\titlespacing{\subsection}{0pt}{1.5ex}{1.2ex}
\titlespacing{\subsubsection}{0pt}{1ex}{0.9ex}
\makeatletter
\renewcommand*{\@seccntformat}[1]{\csname the#1\endcsname .\hspace{0.7em}}
\makeatother

\usepackage{fancyhdr,lastpage}
\fancypagestyle{firststyle}
{
   \fancyhf{}
	\lfoot{IEEE SPMB 2021}\rfoot{v1.0: June 1, 2021}
}
\title{\spmbtitlefont SCORE-IT: A Machine Learning-based Tool for Automatic Standardization of EEG Reports
{\vspace{-2.4\baselineskip}
}
}
    \author{\spmbauthorfont\IEEEauthorblockN{
    Sam Rawal\textsuperscript{\it 1} and 
    Yogatheesan Varatharajah\textsuperscript{\it 2,3} 
    }
    \vspace{0.9em}
    \IEEEauthorblockA{\spmbauthorfont
        1. Carle Illinois College of Medicine, University of Illinois at Urbana Champaign.\\
        2. Department of Bioengineering, University of Illinois at Urbana Champaign.\\
        3. Department of Neurology, Mayo Clinic.\\
        \{scrawal2, varatha2\}@illinois.edu
    }
}

\newcommand{\AbstractSummary}{S.\ Rawal, et al.: SCORE-IT}

\begin{document}

\IEEEaftertitletext{}
\maketitle

\begin{abstract}
Machine learning (ML)-based analysis of electroencephalograms (EEGs) is playing an important role in advancing neurological care. However, the difficulties in automatically extracting useful metadata from clinical records hinder the development of large-scale EEG-based ML models. EEG reports, which are the primary sources of metadata for EEG studies, suffer from lack of standardization. Here we propose a machine learning-based system that automatically extracts components from the SCORE specification from unstructured, natural-language EEG reports. Specifically, our system identifies (1) the type of seizure that was observed in the recording, per physician impression; (2) whether the session recording was \textit{normal} or \textit{abnormal} according to physician impression; (3) whether the patient was diagnosed with epilepsy or not. We performed an evaluation of our system using the publicly available TUH EEG corpus and report F1 scores of 0.92, 0.82, and 0.97 for the respective tasks.



\end{abstract}

\IEEEpeerreviewmaketitle    
\thispagestyle{firststyle}  
\section{Introduction}
\label{sec:intro}
Analysis and interpretation of EEG data is important in the diagnosis of neurological conditions such as epilepsy. Recently, machine learning-based analyses are becoming the mainstay of quantitative EEG-based decision making \cite{golmohammadi2019automatic}. EEG reports are the primary source of metadata for quantitative EEG studies because they contain rich information regarding the patients' condition at the time of EEG recording including seizures, interictal abnormalities, and background activity such as posterior dominant rhythm. However, due to a lack of standardization in the organization of information presented and terminology in EEG reports, automatic extraction of those information has proven to be a difficult task. As a result, manual information retrieval has been the primary way of generating metadata for EEG-based ML studies. Clearly, this is not scalable and warrants the development of natural language processing-based automated tools.

The standardized computer-based organized reporting of EEG (SCORE) guidelines seek to standardize reporting of EEG studies by specifying the content and terminology of characteristics that are described in an EEG report \cite{beniczky2013standardized, beniczky2017standardized}. As such, those guidelines provide an ideal target for automated information retrieval and validation. In this work, we present a semi-automated tool for extracting structured information from unstructured natural text EEG reports with an eye towards converting past EEG reports to the standardized format stipulated in the SCORE guidelines. Furthermore, we also leverage the classification tasks made available through various labeled subsets of the TUH dataset as a valuable proxy to evaluate the capabilities of our system.


The development of such a system poses several challenges: 1) data sparsity: from an entire patient record consisting of dozens of sentences, often only a single phrase is relevant to making a classification decision; 2) lack of labeled data: the lack of clinical data available around many of the SCORE attributes; and 3) accounting for varied clinician practices. We took a two-step approach to address those challenges: first, we leverage a previously developed named entity recognition approach using BERT Transformer models trained on the National NLP Clinical Challenges (n2c2) dataset \cite{patrick2010high}; second, hand-crafted rules are applied to these extracted entities, considering factors such as the SCORE lexicon and numerical values to identify information relevant to SCORE attributes. This hybrid approach allows us to leverage a) existing well-trained ML models for relevant entity identification, and b) domain knowledge to further classify them into SCORE attributes.

In the current work, we focus on identifying three SCORE attributes: (1) the type of seizure if present (complex partial, simple partial, absence, myoclonic, generalized tonic-clonic, other/none); (2) whether the recording contained any abnormal events (e.g., seizures); and 3) whether the patient is being evaluated for epilepsy. In addition, we designed and validated our system on the Temple University Hospital (TUH) EEG dataset \cite{harati2014tuh}, containing over 16,000 EEG recordings and reports. We utilized subsets of the TUH EEG corpus consisting ground truth labels for the three tasks we focused on. We achieved weighted F1 scores of 0.92 (on 171 test records), 0.82 (on 561 records), and 0.97 (on 2727 records) for the seizure classification, normal/abnormal classification, and epilepsy classification tasks, respectively.

\section{Related Work}

Existing work on classification from patient medical records includes Track 1 of the 2018 National NLP Clinical Challenges (n2c2), which involves making binary classification decisions about whether a patient meets some criteria, such as alcoholism or history of myocardial infarction over past 6 months \cite{stubbs2019cohort}. Existing strategies for this type of task include extraction of clinical entities, hand-crafted features that are used to train simple machine learning models, and creation of lexicons to feed into rules \cite{rawal2020semi}. On the other hand, there are related classification tasks in the clinical domain for which there is ample labeled data available, e.g., clinical/biomedical named entity recognition (NER) \cite{patrick2010high}. Prior work on performing NER has included the use of classical Machine Learning models, such as Conditional Random Fields, as well as deep neural networks, including LSTM networks \cite{rawal2018prescription} and, more recently, fine-tuning transformer networks, such as BERT \cite{devlin2018bert}.


However, our work on automatically extracting useful metadata such as seizure type, EEG classification, and diagnostic information from EEG reports, to our knowledge, is the first attempt at developing automated information retrieval approaches for EEG reports. Considering that the majority of the (past and present) EEG reports are written in natural text format, our work offers the potential to standardize EEG reports and to generate useful metadata for subsequent ML-based analyses.

\section{Data}

The TUH EEG Corpus (TUEG) is a collection of over 30,000 clinical EEG records collected and made available by Temple University Hospital (TUH) \cite{harati2014tuh}. Corresponding patient medical reports, in plaintext format, are also made available alongside the EEG recordings. There are also subsets of the corpus, containing labeled data for several tasks. In this work, we utilize the TUH EEG Epilepsy Corpus (TUEP), TUH Abnormal EEG Corpus (TUAB), and TUH EEG Seizure Corpus (TUSZ). Table \ref{table:datatable} describes the number of labeled samples for each of the subsets. (As the Epilepsy subset had no train and test partition, unlabeled records from outside the subset were used to develop rules.)

\begin{table}[h!]
%
\centering
\caption{TUH Seizure Dataset Support}
\label{table:datatable}
 \begin{tabular}{|C{0.15\linewidth}|C{0.25\linewidth}|C{0.12\linewidth}|C{0.12\linewidth}|} 
 \hline
{\GREY Dataset} & {\GREY Class} & {\GREY Train Support} & {\GREY Test Support}  \\
Seizure & Absence & 10 & 6 \\\hline
 & Complex Partial & 45 & 13 \\\hline
 & Myoclonic & 1 & 12 \\\hline
 & Simple Partial & 2 & 0 \\\hline
 & Tonic-Clonic & 12 & 4 \\\hline
 & None & 913 & 97 \\\hline

Epilepsy & Epilepsy    & 428 & 428 \\\hline
 & No Epilepsy & 133 & 133 \\\hline

Abnormal & Normal    & 1371 & 150 \\\hline
 & Abnormal & 1346 & 126 \\\hline

\end{tabular}
\end{table}
    

\section{Methodology}
In this work, we present a system that, given a natural-language patient report, predicts, 1) the type of seizure that was observed in the recording, per physician impression; 2) whether the session recording was \textit{normal} or \textit{abnormal} according to physician impression; and 3) whether a patient has epilepsy or not.

To construct such a system, we provide a framework for classifying EEG records that divides the process into two steps: a \textit{broad parsing} step, for which tasks are well-defined and considerable training data exists; and a \textit{narrow parsing/classification} step built on top of the broad parsing, for which tasks are highly domain-specific and little training data is available. In this manner, we address issues of data sparsity and support while simultaneously leveraging effective data-driven, deep learning techniques that can be adapted to this specific domain.

\begin{figure}[!h]
\vspace{0.5em} 
    \centering
    \captionsetup{justification=centering}
    \includegraphics[width=0.9\linewidth]{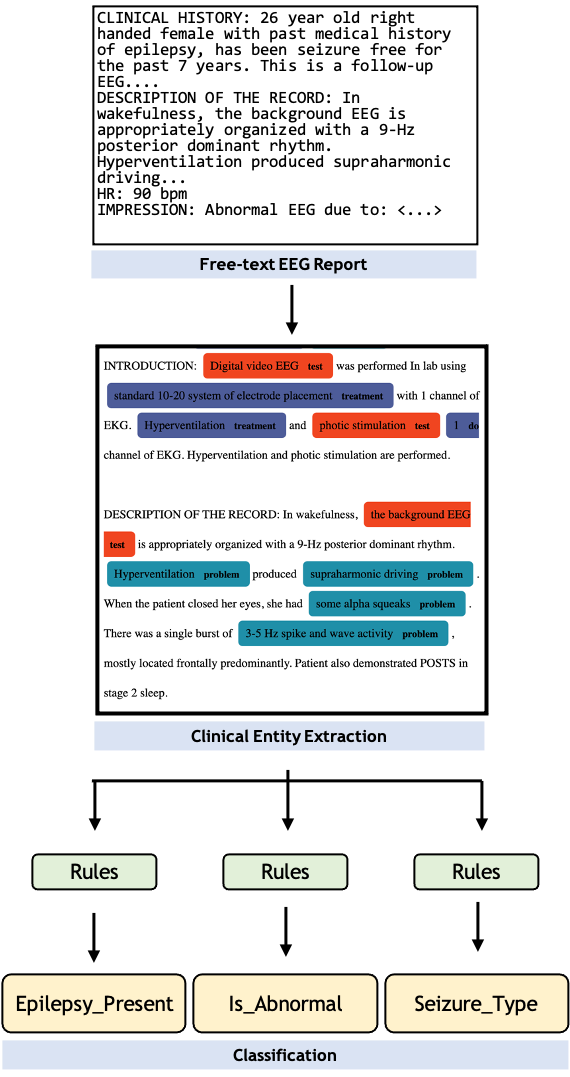}
    \caption{System Architecture.}
    \vspace{-0.75em} 
    \label{fig:architecture}
\end{figure}

A high-level architecture of the system is detailed in Figure \ref{fig:architecture}. In the following subsections, we discuss each component.

\subsection{Broad Parsing}
The overall goal of this step is to reduce the data sparsity by only extracting information from the record that will be potentially useful for classification, while leveraging existing well-trained machine learning models and methods that utilize the structure of the medical records.

First, section headers are identified, and each sentence in the report is matched with its corresponding header. The headers are extracted using a regular expression that matches the format of the medical records present in the TUH corpus. Second, we perform Named Entity Recognition using BERT Transformer models \cite{devlin2018bert} trained on datasets released by national NLP clinical challenges (n2c2). Specifically, we extract clinical entities, such as medical problems, labs, and treatments, and medication entities, such as medication name, dose, frequency, duration, and reason. The information extracted in this step is then passed to the \textit{narrow parsing} step.

\subsection{Narrow Parsing}
The \textit{narrow parsing} step that is built on top of the broad parsing step. This step is motivated by the fact that the classification tasks we address have a small support of training data; therefore, it is not possible to train a system only from the supervised data. Consequently, this phase consists of a series of hand-crafted rules built around the extracted entities and sections in the prior step, in effect performing domain adaptation from models trained on related tasks in the broad parsing step to these three tasks with limited data.

Classification of each of the of three tasks (epilepsy, normal/abnormal, seizure type) is performed using hand-crafted rules that are constructed using the outputs of the section and entity extraction from the previous step. Thus, the broad parsing step dramatically compresses the input dimensionality from an unstructured, free-text medical report to a collection of named entities and the section of the note they are identified in, thus making it possible to create simpler and more generalizable rules. 

For example, the rule used for binary classification in the Epilepsy identification task consists of extracting all \textit{problem} entities mentioned in the ``CLINICAL HISTORY" section of the patient record, then outputting a positive classification if any of the entities match \textit{epilepsy} (or synonyms and related phrases).

\section{Experimental Design}
The system was evaluated on three specific tasks that are subsets of the TUH EEG dataset, and for which gold annotations are provided: 
\begin{enumerate}
    \item Seizure type classification: given a record, determine the type of seizure noted in the EEG record (complex partial, simple partial, absence, myoclonic, tonic-clonic, other/none).
    \\
    \item Epilepsy classification: given a record, determine whether the patient is being evaluated for epilepsy or not.
    \\
    \item Norma/abnormal classification: determine if the clinical impression in a given record is \textit{normal} or \textit{abnormal}.
\end{enumerate}

A detailed information for each of these tasks is provided in the following subsections.

\subsection{Seizure Classification}
The Seizure subset of the TUH dataset contains 1010 unique patient records classified by the type of seizure that is presented in the corresponding EEG recording. One important aspect to note is that, in this work, we focused on specific types of seizures or NONE. Consequently, records labeled ``GNSZ" or ``FNSZ" (for generalized or focal seizures respectively, without any further specifics) were dropped from the dataset. The breakdown of support for each type of seizures is described in Table \ref{table:datatable}. 

\subsection{Epilepsy Classification}
The Epilepsy subset of the TUH dataset contains 428 patient records labeled \textit{epilepsy} and 133 records labeled \textit{no epilepsy}. One complication of this dataset is that the labels are associated with the EEG recording findings, rather than what is detailed int he patient records; in a manual review of the dataset, it appears there are numerous instances of the gold label not being substantiated within the clinical record alone. 

\subsection{Normal/Abnormal Classification}
The Abnormal subset of the TUH corpus consists of notes that are labeled \textit{normal} or \textit{abnormal}. This classification is made based off physician comments in the ``CLINICAL IMPRESSIONS" section of medical records. There are 126 notes with the \textit{abnormal} label and 150 notes with the \textit{normal} label in this subset.

\section{Results}
On the seizure classification task, our system performed with a weighted F1 score of 0.92 on 171 test records. On the epilepsy classification task, our system performed with a weighted F1 score of 0.82 on 561 records. On the abnormal classification task, our system performed with a weighted F1 of 0.97 on 2727 records. Notably, the system performed poorer on classes with a very small support, due to difficulty in creating generalizable hand-crafted rules from a very small dataset. More detailed results are provided in Table \ref{table:results}.

\begin{table}[!h]
%
\centering
\caption{Classification Results -- Weighted Average}
\label{table:results}
 \begin{tabular}{|C{0.15\linewidth}|C{0.15\linewidth}|C{0.12\linewidth}|C{0.12\linewidth}|C{0.12\linewidth}|} 
 \hline
{\GREY Task} & {\GREY Precision} & {\GREY Recall} & {\GREY F1-score} & {\GREY Support} \\
Seizure & 0.93 & 0.93 & 0.93 & 121 \\\hline
Abnormal & 0.98 & 0.97 & 0.97 & 276 \\\hline
Epilepsy & 0.82 & 0.82 & 0.82 & 561 \\\hline
\end{tabular}
\end{table}

\section{Discussion}
We proposed an automated approach for extracting clinically useful metadata from natural text EEG reports. Major challenges addressed in our work include 1) designing strategies for integration of clinical domain knowledge, 2) accounting for varied clinician practices, 3) establishing ground truth standards, and 4) robust verification of the system. Applications of this system include enabling better search and filtering of clinical reports, as well as auto-labeling unstructured datasets for research purposes.

As mentioned previously, one complication with interpreting these results is that for certain tasks, such as epilepsy classification, the labels are derived from the EEG findings and not necessarily explicitly mentioned in the patient medical record, making it difficult to fully accurately judge the performance of the system. 

Although the broad and narrow parsing architecture of the system helped in the development of more general rules, the system still performed better on classes with greater support. This can be attributed to the fact that having access to a wider range of training samples allowed rules to be developed that could capture more of the variance. On the other hand, the system performed poorer on classes with few samples, as the rules developed with under 5 training samples were not able to be validated to ensure sufficient generalizability.

\section{Future Work}
The current iteration of the system still performs relatively poorly in classification circumstances with low data support. Consequently, one important area of focus is reducing the amount of hand-crafted rules needed, while increasing the generalizability of the system, in order to be able to apply the system across a wider range of use-cases. To that effect, a prominent area of future focus remains using existing NLP tasks, such as semantic textual similarity \cite{wang20202019sts}, as a means to perform zero-shot classification in place of hand-crafting rules. Another area for future work is implementing and validating our approach on additional datasets.

\section{Conclusion}

We described a semi-automated ML-based system to extract standardized components from unstructured EEG reports. Our system can facilitate better indexing, searching and organization of existing EEG report collections by providing an semi-automated methodology of standardizing important data within free-text EEG reports. Furthermore, our system can be used to generate labels for large, unlabeled EEG corpora like the TUH dataset to motivate future research.



\section*{Acknowledgements}
This research was supported by National Science Foundation's Computer and Information Science and Engineering Research Initiation Initiative Award SCH-2105233. We would like to thank Dr. Susan Herman and her team from the Barrow Neurological Institute, as well as Neeraj Wagh from the University of Illinois at Urbana Champaign, for contributions to discussions on where this work can be applied.

\footnotesize
\bibliographystyle{IEEEbibSPMB}
\bibliography{IEEEabrv,IEEESPMB}

\end{document}